\documentclass{article}
\usepackage[T1]{fontenc}
\usepackage{subcaption}
\usepackage{booktabs}
\usepackage{hyperref}
\usepackage{orcidlink}      
\usepackage{authblk}        
\usepackage{tabularx}
\usepackage{array}
\usepackage{xurl}
\usepackage{ragged2e}
\usepackage{makecell}
\usepackage{float}
\usepackage{placeins}
\usepackage{graphicx}
\usepackage{xcolor}
\usepackage{appendix}
\usepackage{threeparttable}  
 
\title{Unpacking Vibe Coding: Help-Seeking Processes in Student-AI Interactions While Programming}
 
\author[1]{Daiana Rinja\,\orcidlink{0009-0002-1461-4727}}
\author[2]{Eduardo Araujo Oliveira\,\orcidlink{0000-0001-5063-8860}}
\author[3]{Sonsoles López-Pernas\,\orcidlink{0000-0002-9621-1392}}
\author[3]{Mohammed Saqr\,\orcidlink{0000-0001-5881-3109}}
\author[1]{Marcus Specht\,\orcidlink{0000-0002-6086-8480}}
\author[1]{Kamila Misiejuk\,\orcidlink{0000-0003-0761-8703}}
 
\affil[1]{CATALPA, FernUniversität in Hagen, 58097 Hagen, Germany\\ \texttt{\{daiana.rinja, marcus.specht, kamila.misiejuk\}@fernuni-hagen.de}}
\affil[2]{School of Computing and Information Systems, University of Melbourne, Parkville VIC 3010, Australia\\ \texttt{eduardo.oliveira@unimelb.edu.au}}
\affil[3]{School of Computing, University of Eastern Finland, 80110 Joensuu, Finland\\ \texttt{\{sonsoles.lopez, mohammed.saqr\}@uef.fi}}
 
\date{}
 
\begin{document}
\maketitle

\begin{abstract}
Generative AI is reshaping higher education programming through vibe coding, where students collaborate with AI via natural language rather than writing code line-by-line. We conceptualize this practice as help-seeking, analyzing 19,418 interaction turns from 110 undergraduate students. Using inductive coding and Heterogeneous Transition Network Analysis, we examined interaction sequences to compare top- and low-performing students. Results reveal that top performers engaged in instrumental help-seeking -- inquiry and exploration -- eliciting tutor-like AI responses. In contrast, low performers relied on executive help-seeking, frequently delegating tasks and prompting the AI to assume an executor role focused on ready-made solutions. These findings indicate that currently generative AI mirrors student intent (whether productive or passive) rather than optimizing for learning. To evolve from tools to teammates, AI systems must move beyond passive compliance. We argue for pedagogically aligned design that detect unproductive delegation and adaptively steer educational interactions toward inquiry, ensuring student-AI partnerships augment rather than replace cognitive effort.
 
\end{abstract}
 

%
%
\section{Introduction}

Generative AI (AI) has enabled a new mode of programming described as \textit{vibe coding}: a shift from writing code line-by-line toward an interaction-driven workflow in which programmers iteratively specify their intentions in natural language, evaluate generated outputs, and refine their requests. In this mode, AI expands the role of a reference tool to become an active participant in the development process, akin to a teammate that proposes solutions, diagnoses issues, and suggests next steps \cite{feng2024coprompt,penney2025outcomes}. While such workflows can reduce friction and accelerate development for experienced programmers, for novice programmers they also represent a new way of seeking help -- one in which computational thinking is mediated by conversational problem solving.

This shift in programming practice has rapidly permeated higher education, where students have begun to incorporate AI chat assistants such as ChatGPT into their everyday programming work, frequently using them as their primary source of support during coding tasks. Early evidence suggests that many learners turn to AI before consulting instructors, peers, or static resources, with some reporting AI as their first resort of help-seeking \cite{hyrynsalmi2025person}. However, this widespread reliance on on-demand assistance introduces a critical tension between efficiency and educational value. Over-reliance on delegation to an AI assistant may short-circuit essential learning processes such as productive struggle, debugging, and self-explanation \cite{chen2025unpacking,fan2025beware}.

A growing body of research has examined how students use AI in programming. Existing studies, however, have focused on usage patterns, aggregate metrics, and student perceptions of benefits and risks. While recent work has started to adopt process-oriented perspectives \cite{fan2025beware,chen2025unpacking}, most studies treat the interaction as a ``black box'', focusing on the final code artifact or the student's sentiment rather than the dialogue itself. We lack fine-grained evidence that characterizes student-AI collaboration \textit{as an unfolding interactional process}. Specifically, it remains unclear how students initiate help-seeking, how the AI responds, and the granular sequences through which collaboration evolves.

To address this gap, we analyze a large dataset of authentic student-AI interactions collected during an undergraduate artificial intelligence course.  By examining the temporal dynamics of these dialogues, we investigate the following research questions:
\textbf{RQ1}: How do student–AI interaction patterns during programming tasks differ between top- and low-performing students? \textbf{RQ2}: How do student help-seeking strategies elicit AI-enacted roles during programming between top- and low-performing students? \textbf{RQ3}: To what extent do student-AI interaction patterns during programming tasks predict student performance?

\section{Background}

\subsection{Help-Seeking in Programming}

Help-seeking is a key metacognitive strategy within self-regulated learning theory, serving as an adaptive behavior that supports learning when students encounter challenges beyond their current capabilities \cite{pintrich2004conceptual}. Rather than signaling weakness, effective help-seeking reflects goal-directed decision-making that enables learners to regulate their progress and overcome obstacles strategically \cite{karabenick2013help}. A fundamental distinction in the literature is between \textit{instrumental} and \textit{executive} help-seeking \cite{karabenick2013help}. Instrumental help-seeking involves asking for hints or guidance to develop the knowledge needed to solve a problem and reduce future dependence. In contrast, executive help-seeking refers to requesting direct answers to finish a task with minimal cognitive engagement. Instrumental help-seeking is associated with academic success, unlike executive help-seeking, which often correlates with shallower learning outcomes \cite{aleven2003help}.

This distinction is particularly critical in programming, a domain where students frequently face multilayered difficulties, ranging from syntax errors to complex design and decomposition challenges \cite{ettles2018common}. To address these obstacles, students traditionally navigated a diverse range of resources such as instructors, teaching assistants, peers, and most notably online communities like Stack Overflow. These platforms have fostered a culture of shared knowledge, where help-seeking required learners to precisely articulate problems and engage in peer review \cite{staegemann2025did}. However, as learners pivot from public, community-based inquiry to private, instant AI interactions, empirical studies indicate a marked decline in forum participation \cite{staegemann2025did}. While this shift offers efficiency, it fundamentally alters the learning journey: the ``productive struggle'' of formulating a public question is replaced by dyadic delegation, potentially diminishing the cognitive depth associated with voluntary knowledge contribution \cite{shan2025examining}.

Thus, AI introduces an interactional mode of help-seeking distinct from traditional resources. In the AI-assisted workflows described as vibe coding, learners externalize parts of planning, implementation, and debugging into repeated conversational requests, effectively embedding help-seeking into the moment-to-moment flow of programming \cite{denny2024computing}. However, a critical tension arises in how learners process this support. While human tutoring involves clarification and negotiation \cite{karabenick2013help}, students interacting with AI may implement suggestions quickly with limited reflection or evaluation \cite{chen2025unpacking}. This creates a risk that AI convenience reinforces executive help-seeking patterns at the expense of the deeper engagement found in traditional instruction \cite{kazemitabaar2023studying}.

\subsection{Student-AI Collaboration in Programming}

Given AI's expansion, a substantial strand of research has examined the outcomes of integrating AI tools into programming learning contexts \cite{Nathaniel2025-vl}. 
Results have been mixed: Cubillos et al. \cite{Cubillos2025-tm} found no differences in learning outcomes between AI support and instructional videos, while Garg et al. \cite{Garg2025-oy} demonstrated that AI alone did not guarantee better outcomes, though structured prompt training significantly improved performance. 
To further understand these mixed results, a growing body of literature investigates interaction processes to reveal \textit{how} students engage with these tools. 
Amoozadeh et al. \cite{Amoozadeh2024-od} examined student–AI collaboration in introductory programming by observing how students used ChatGPT while solving Python tasks. Their study showed that students often relied heavily on the tool, directly submitting the full task descriptions without prior individual effort, and conducted only limited verification of the generated solutions. López-Pernas et al. \cite{Lopez-Pernas2025-gt} studied how students interact with AI during programming tasks by analyzing their prompts over time, shifting the focus from learning outcomes to interaction processes. Their findings showed that students predominantly used AI for monitoring progress and solving immediate problems, while rarely engaging in deeper metacognition such as reflection or evaluation. This aligns with observations of ``premature convergence'' in vibe coding, where learners accept the first viable solution generated by the AI without critical iteration or alternative exploration \cite{gama2025can}. 

We argue that these behaviors stem from the specific roles enacted during the interaction. Treating student-AI interactions as a series of isolated prompts overlooks that AI responses are shaped turn-by-turn and can simulate different pedagogical roles, from a \textit{tutor} offering scaffolded hints to an \textit{executor} providing direct code. These roles are enacted through how students distribute agency across the interaction. Learners now act as orchestrators of AI tool pipelines, shifting their focus from syntax generation to intent specification \cite{gama2025can}. This shift raises critical questions about when students delegate goal-setting to the AI versus when they retain control themselves \cite{fan2025impact,prather2024widening}.

This presents a critical gap in our understanding of AI-mediated programming practice. Even studies that analyze interaction data typically focus on aggregate usage patterns or isolated prompt classifications, rather than on the sequential and process dynamics of student–AI exchanges \cite{Amoozadeh2024-od,wang2023matters,mailach2025ok}. We lack research that examines the dialogic nature of these interactions, where meaning and solutions are co-constructed across multiple turns. To address this gap, this study investigates \textit{vibe coding} in an authentic university course. We conceptualize student-AI collaboration as a dynamic, multi-turn process shaped by the interplay between help-seeking strategies and enacted AI roles. By moving beyond aggregate metrics to analyze the granular sequencing of these interactions, we aim to unpack how different collaborative dynamics (specifically those top- and low-performers) facilitate or hinder learning outcomes.
\section{Methods}

\subsection{Dataset}

We use the StudyChat dataset \cite{mcnichols2025studychat}, publicly available on HuggingFace (\href{https://huggingface.co/datasets/wmcnicho/StudyChat}{huggingface.co/datasets/wmcnicho/StudyChat}). The dataset contains student prompts and responses generated by a web-based AI assistant powered by GPT-4o-mini, which students were encouraged to use freely across course assignments in an undergraduate artificial intelligence course (COMPSCI 383 at the University of Massachusetts Amherst). The data were collected during the Spring 2025 academic year and include interaction logs, individual assignment grades, as well as grades from two end-of-module exams and a final exam from consenting students \cite{mcnichols2025studychat}. The course comprised seven assignments spanning a variety of activities, including video-based reflections, quiz questions, code generation, output evaluation, and written explanations of code intuition. Programming tasks were delivered through Jupyter notebooks with progressively reduced scaffolding over time, and several assignment components explicitly encouraged the use of the chatbot to support task completion. Overall, the dataset includes 9,709 student prompts and 9,709 chatbot responses (n = 19,418 interactions) from 110 students across 1,213 chats.

\subsection{Coding of Student-AI Interactions}

We developed two complementary coding schemes: one to map the elements of student prompts and AI responses (\textit{interaction elements}), and another to categorize student prompt types and AI roles (\textit{interaction types}). This approach allowed us to analyze student–AI interactions at both a granular level (detailed elements) and a higher level (overall interaction patterns). In both schemes, the unit of analysis is the individual message: each student prompt or AI response constitutes one codable unit.

The coding scheme for interaction elements was developed through an inductive coding process (see Table \ref{table:coding_scheme}). Two researchers divided the coding task by role: one coded all student prompts and the other coded all AI responses. Each researcher began by independently coding an initial subset, after which emerging codes were discussed and differences were reconciled. This process was repeated over multiple rounds, with each iteration refining code definitions, particularly the boundary cases requiring clarity. A single message could receive multiple codes, applied in order of which the elements appeared within it. Once the codebook was finalized, each researcher coded their full assigned subset. To assess inter-rater reliability, the researchers cross-coded 10\% of the dataset ($n = 2,000$ interactions). Because a single message could receive multiple element codes, we computed Cohen's $\kappa$ separately for each code by treating it as a binary decision (present vs. absent) at the message level. Interrater reliability was at least substantial for both student prompts (per-code $\kappa = 0.906$--$0.961$) and AI responses (per-code $\kappa = 0.738$--$0.929$). After establishing sufficient reliability, the remaining dataset was divided and coded by individual researchers.

Building on the interaction element coding, we developed a rule-based scheme to aggregate the 34 unique element combinations observed in student prompts and 18 in AI responses into broader interaction types, each assigned once per message. Prompt types were determined by the presence or absence of key elements: prompts containing \textit{exploration} without any task-specific context (\textit{code}, \textit{assignment}, \textit{results} or \textit{error}) were classified as \textit{Inquire}, while those combining \textit{exploration} with at least one contextual element were classified as \textit{Integrate}. Prompts containing \textit{error} without \textit{exploration} were classified as \textit{Debug}, and those containing task-oriented elements with neither \textit{exploration}, nor \textit{error}, were classified as \textit{Delegate}. AI roles followed analogous logic: responses containing \textit{solution} without \textit{feedback} were classified as \textit{Executor}, those containing both as \textit{Collaborator}, those with \textit{feedback} but no \textit{solution} as \textit{Evaluator}, and those containing only \textit{explanation}, \textit{instruction}, and/or \textit{example} as \textit{Tutor}. Each student prompt or AI response was assigned to exactly one interaction type. This aggregation offers a higher-level overview of help-seeking strategies and AI roles while preserving clear interpretability.

Prompt types reflect either instrumental help-seeking aimed at understanding and conceptual clarification (\textit{Inquire}, \textit{Integrate}) or executive help-seeking that delegates task completion or requests direct solutions (\textit{Debug}, \textit{Delegate}) \cite{karabenick2013help}. All AI roles except \textit{Executor} function as pedagogical agents, providing guidance or feedback. In contrast, the \textit{Executor} role reflects a non-pedagogical mode delivering ready-made solutions without guidance. This distinction is consistent with prior work on pedagogical agents and intelligent tutoring systems, which emphasizes explanation, feedback, and scaffolding as core features of pedagogically meaningful support \cite{denny2024computing}.

\subsection{Analysis}
To contrast performance extremes (RQ1, RQ2), we stratified students by grade quartiles based on a composite score of assignments and exams, selecting: Q1 (top performance; mean grade = 0.96, n = 27 students, 5,420 interactions) and Q4 (low performance; mean grade = 0.86, n = 28 students, 2,946 interactions). All data analyses were conducted in R using the \texttt{TNA} package \cite{tikka2025tna}, \texttt{tnaExtras} and \texttt{dplyr}. Additional graphs and tables showing the analysis across all performance groups are available in the \hyperref[sec:appendix]{Appendix}.

We employed Heterogeneous Transition Network Analysis (HTNA) \cite{htna2026} to examine student–AI interaction sequences. HTNA  extends Transition Network Analysis \cite{saqr2025transition} to heterogeneous networks, which are composed of two or more distinct node types connected by edges that can span across or within node types \cite{sun2013mining}. Unlike bipartite approaches, HTNA preserves both within-type and cross-type transitions in a single network, retaining the full sequential structure of the interaction data \cite{htna2026} using first-order Markov models.

We developed two HTNA models to compare transitions among interaction elements (RQ1) and interaction types (RQ2). In both models, the two node types correspond to student and AI categories. Because each message may contain multiple elements applied in sequence, the resulting interaction chain includes both within-type transitions (e.g., \textit{request}$\rightarrow$\textit{code} within a single student prompt) and cross-type transitions (e.g., \textit{code}$\rightarrow$\textit{solution} from student to AI). We compared edge strengths by calculating the difference between edge weights in the Q1 and Q4 networks using permutation tests to identify statistically significant edges. In addition, we employed Pearson residual analysis with permutation tests to assess differences in individual code frequencies (RQ1, RQ2) and chi-squared tests to compare sequence proportion distributions (RQ1, RQ3). To address RQ3, we conducted two linear regression models using average student grade as the dependent variable predicted by the proportion of interaction elements and interaction types per student. Both models included all student data and removed variables with high multicollinearity.

\begin{table}[h!]
\centering
\scriptsize
\caption{Coding schemes for student-AI interactions}
\begin{tabular}{
    >{\raggedright\arraybackslash}p{1.8cm}
    >{\raggedright\arraybackslash}p{1cm}
    >{\raggedright\arraybackslash}p{9.2cm}
}
\toprule
\textbf{Code} & \textbf{Freq} & \textbf{Description} \\
\midrule

\multicolumn{3}{l}{\textbf{Interaction elements: \textit{Student prompts}}} \\

Assignment &2,261 
& Student pastes instructions from an assignment. 
 \\
 
Code  &1,634 & Student pastes code directly. 
 \\

 Error &698 
& Student provides an error message or describes an error.
\\ 

Exploration & 3,567 
& Student asks questions, seeks clarification or feedback. \\

Request  &2,768 
& Student instructs the AI to perform a task or write code.  \\

Results   & 972 
& Student provides raw output, results, or data. 
 \\

\midrule

\multicolumn{3}{l}{\textbf{Interaction elements: \textit{AI responses}}} \\
Feedback  &2,095 
& AI addresses the student's code, output, or error directly. \\

Explanation  &6,722 
& AI gives conceptual explanation and/or explains key steps. \\

Instruction &2,710 
&  AI provides explicit procedural guidance. \\

Solution  &5,192
&  AI generates a ready-to-use fix or answer. \\

Example &3,825
&  AI shows general or supporting examples. \\

\midrule

\multicolumn{3}{@{}l}{\textbf{Interaction types: \textit{Prompt types}}}\\
Inquire &2,571 
& Conceptual inquiry focused on understanding or clarification (contains \textit{Exploration} and no \textit{Code, Assignment, Results, or Error}.) \\

Integrate &911 
& Applied inquiry that combines conceptual exploration with task-specific context (contains \textit{Exploration} and at least one of: \textit{Code, Assignment, Results, or Error}). \\

Debug &600 
& Focuses on fixing a problem in existing work (contains \textit{Error}, no \textit{Exploration}). \\

Delegate&5,504
& Focuses on delegating execution of a task to the AI
(contains \textit{Request, Code, Assignment, or Results}, with no \textit{Exploration} and no \textit{Error}). \\

\midrule
\multicolumn{3}{@{}l}{\textbf{Interaction types: \textit{AI roles}}} \\
Executor & 3,228
& Carries out the requested task without engaging with student work (contains \textit{Solution} and no \textit{Feedback}). \\

Collaborator &1,964 
& Provides applied support by responding to student work and contributing to problem solving (contains both \textit{Feedback} and \textit{Solution}). \\

Evaluator &131 
& Focuses on assessing or commenting on student work without completing the task (contains \textit{Feedback} and no \textit{Solution}). \\

Tutor &4,198
& Provides conceptual or procedural guidance without solving or evaluating the task (contains \textit{Explanation}, \textit{Instruction}, and/or \textit{Example}, with no \textit{Feedback} and no \textit{Solution}). \\
\bottomrule
\end{tabular}
\label{table:coding_scheme}
\end{table}

\section{Results}
\subsection{Student-AI Interaction Elements Analysis}
\begin{figure}[h!]
   \vspace{-2mm}
    \centering
    \includegraphics[width=1.0\textwidth]{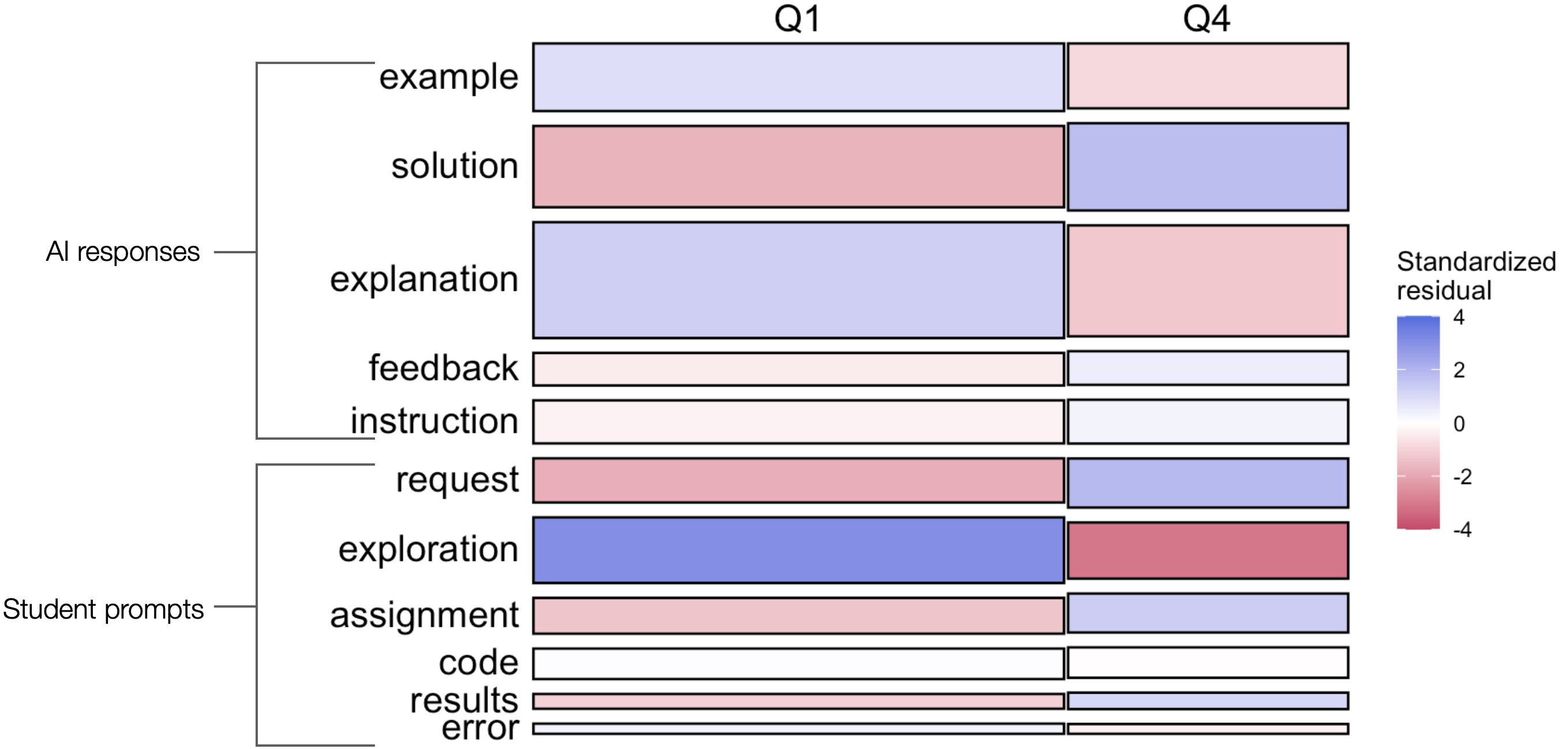}
    \caption{Comparing frequency of individual interaction elements between Q1 and Q4 students. Blue shading indicates overrepresentation, while red shading indicates underrepresentation of each element within a group.}
    \label{fig:mosaic_interaction_element}
\end{figure}

To examine differences in interaction elements within student prompts and AI responses between top-performing (Q1) and low-performing (Q4) students (RQ1), we conducted Pearson residual analysis with permutation testing. The chi-square test revealed that the distribution of individual codes differed significantly between the two groups ($x^2(10) = 18.9, p = 0.04$) (see Fig. \ref{fig:mosaic_interaction_element}). A significant difference was observed in the frequency of the exploration code, which was overrepresented among top-performing students and underrepresented among the low-performing students. Other significant but smaller differences included  the overrepresentation of \textit{example} and \textit{explanation} among Q1 students, and the overrepresentation of \textit{solution}, \textit{request}, \textit{assignment}, and \textit{results} among Q4 students.

\begin{figure}[h!]
    \centering
    \includegraphics[width=0.8\textwidth]{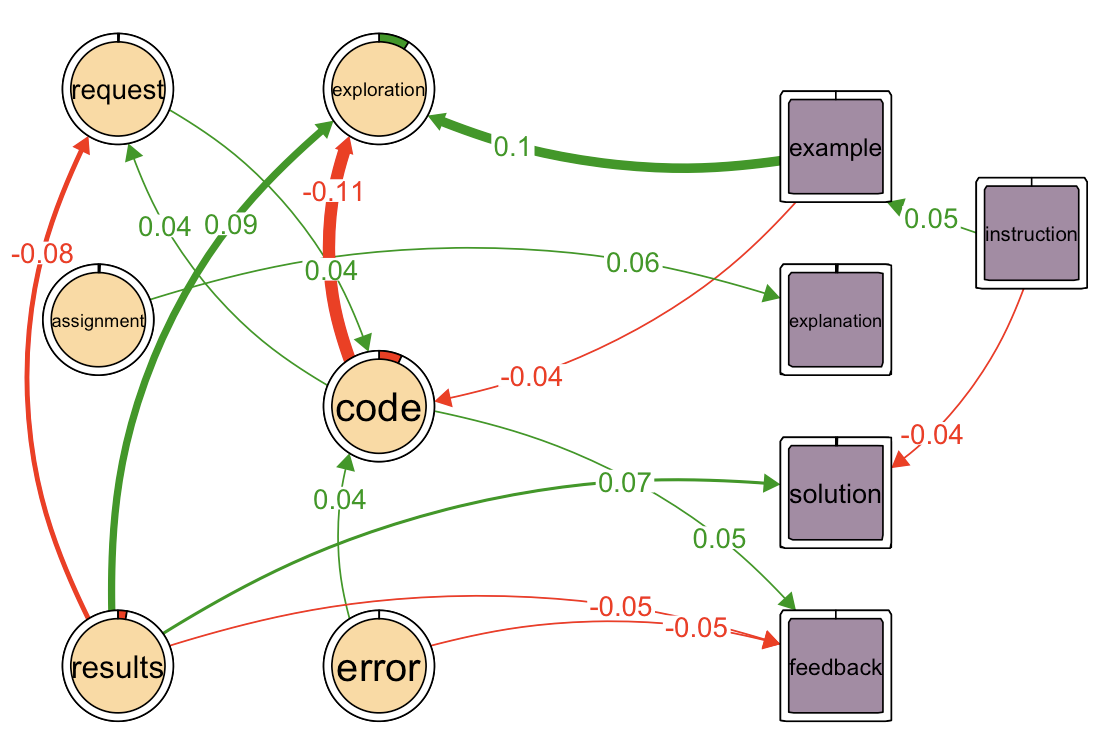}
    \caption{HTNA comparison model for interaction elements. For Q1 students, stronger connections are highlighted in red, whereas for Q4 students, they are highlighted in green. Violet nodes: elements of AI responses; yellow nodes: elements of student prompts. Only transitions over 0.4 are shown.}
    \label{fig:htna_interaction_element}
\end{figure}

The initial probabilities of starting the interaction reveal distinct behavioral differences between Q1 and Q4 students (see Fig.~\ref{fig:htna_interaction_element}). Q4 students were more likely to initiate chats by providing the output of their code or data (\textit{results}) or sharing their \textit{code}, whereas Q1 students posed exploratory questions (\textit{exploration}). Several transitions differentiated the two groups. The most prominent difference in the Q4 group was the transition from \textit{code} to \textit{exploration} (diff: -0.112, $p=0.011$), suggesting students struggled to understand concepts within their own code.
In contrast, Q1 students' stronger transitions to AI \textit{feedback} (diff: 0.05) and a bidirectional loop with \textit{request} indicate that student-written code was either preceded or followed by specific instructions to the AI (\textit{request}$\rightarrow$\textit{code}, diff: 0.04, $p = 0.043$; \textit{code}$\rightarrow$\textit{request}, diff: 0.04). 

Furthermore, Q1 students exhibited significantly stronger transitions from AI-provided \textit{examples} to further \textit{exploration} in subsequent prompts (diff: 0.098, $p=0.019$). In contrast, Q4 students were more likely to transition from \textit{example} to sharing their own \textit{code}, suggesting attempts to implement the example themselves (diff: -0.04). In parts of the assignment, students were explicitly instructed to use AI for tasks such as formatting results. This context explains the following patterns: Q4 students provided more specific instructions through \textit{request} (diff: -0.08), while Q1 students received \textit{solutions} (diff: 0.07). 

Additionally, students used AI to interpret code output (results), revealing another difference: Q1 students were more likely to prompt additional exploration (diff: 0.09), whereas Q4 students engaged with AI to request \textit{feedback} (diff: -0.05). Finally, several differences emerged in AI responses to students of different performance levels. When AI provided step-by-step \textit{instructions} for solving issues, these were more likely to lead to non-assignment related \textit{examples} for Q1 students (diff: 0.05) and to ready-made \textit{solutions} for Q4 students (diff: -0.04). Possible adaptation to student performance level was also evident in the stronger \textit{assignment}--\textit{explanation} transition, where Q1 students received conceptual explanations even after pasting assignment text into the chat (diff: 0.06).

The analysis of differences in longer interaction patterns revealed that Q4 students exhibited repetitive assignment$\rightarrow$\textit{solution} loops, where they pasted assignments and received direct solutions, sometimes repeatedly (see Table \ref{tab:pattern_interaction_element}). Overall, their patterns frequently culminated with AI providing \textit{solutions}, often after minimal intermediate steps. In stark contrast, Q1 students' top patterns were dominated by exploration, appearing in 7 of their 10 most frequent sequences. Their interactions formed iterative learning cycles where AI-provided \textit{examples} and \textit{explanations} trigger further exploratory questions (e.g., \textit{explanation}$\rightarrow$\textit{example}$\rightarrow$exploration). Q1 students also engaged conceptual learning cycles through exploration$\rightarrow$code$\rightarrow$\textit{feedback} sequences, which suggests that they actively worked through problems rather than seeking immediate answers. 

\begin{table}[H]
\centering
\scriptsize
\caption{Top 10 interaction element patterns showing significant frequency differences between Q1 and Q4 students (all p < 0.001). Student prompt elements appear in normal font; AI response elements appear in \textit{italics}.}

\label{tab:pattern_interaction_element}
\begin{tabular}{p{7.1cm} p{0.5cm} p{1.3cm} p{0.5cm} p{1.3cm} p{0.7cm}}
\toprule
Pattern & n & Q1 prop. & n & Q4 prop. & stat. \\
\midrule
\textit{explanation}$\rightarrow$\textit{example}$\rightarrow$exploration & 287 & 0.034 & 103 & 0.023 & 9.78 \\
\textit{example}$\rightarrow$exploration$\rightarrow$\textit{explanation}$\rightarrow$\textit{example} & 192 & 0.023 & 62 & 0.015 & 9.66 \\
exploration$\rightarrow$\textit{explanation}$\rightarrow$\textit{example}$\rightarrow$exploration & 186 & 0.023 & 55 & 0.013 & 12.55 \\
exploration$\rightarrow$code$\rightarrow$\textit{feedback} & 67 & 0.008 & 12 & 0.003 & 11.75 \\
code$\rightarrow$exploration$\rightarrow$\textit{feedback}$\rightarrow$\textit{solution} & 18 & 0.002 & 26 & 0.006 & 11.28 \\
\textit{example}$\rightarrow$request$\rightarrow$\textit{solution}$\rightarrow$\textit{explanation}& 18 & 0.002 & 24 & 0.006 & 9.05 \\
assignment$\rightarrow$\textit{solution}$\rightarrow$assignment & 17 & 0.002 & 33 & 0.007 & 21.03 \\
code$\rightarrow$exploration$\rightarrow$\textit{feedback}$\rightarrow$\textit{solution}$\rightarrow$\textit{explanation}& 12 & 0.002 & 22 & 0.005 & 13.28 \\
request$\rightarrow$\textit{explanation}$\rightarrow$\textit{example}$\rightarrow$code & 12 & 0.001 & 19 & 0.004 & 9.14 \\
assignment$\rightarrow$\textit{solution}$\rightarrow$assignment$\rightarrow$\textit{solution} & 10 & 0.001 & 24 & 0.006 & 18.70 \\
\bottomrule
\end{tabular}
\end{table}

\subsection{Student-AI Interaction Type Analysis}
\begin{figure}[!htbp]
    \centering
    \includegraphics[width=0.9\textwidth]{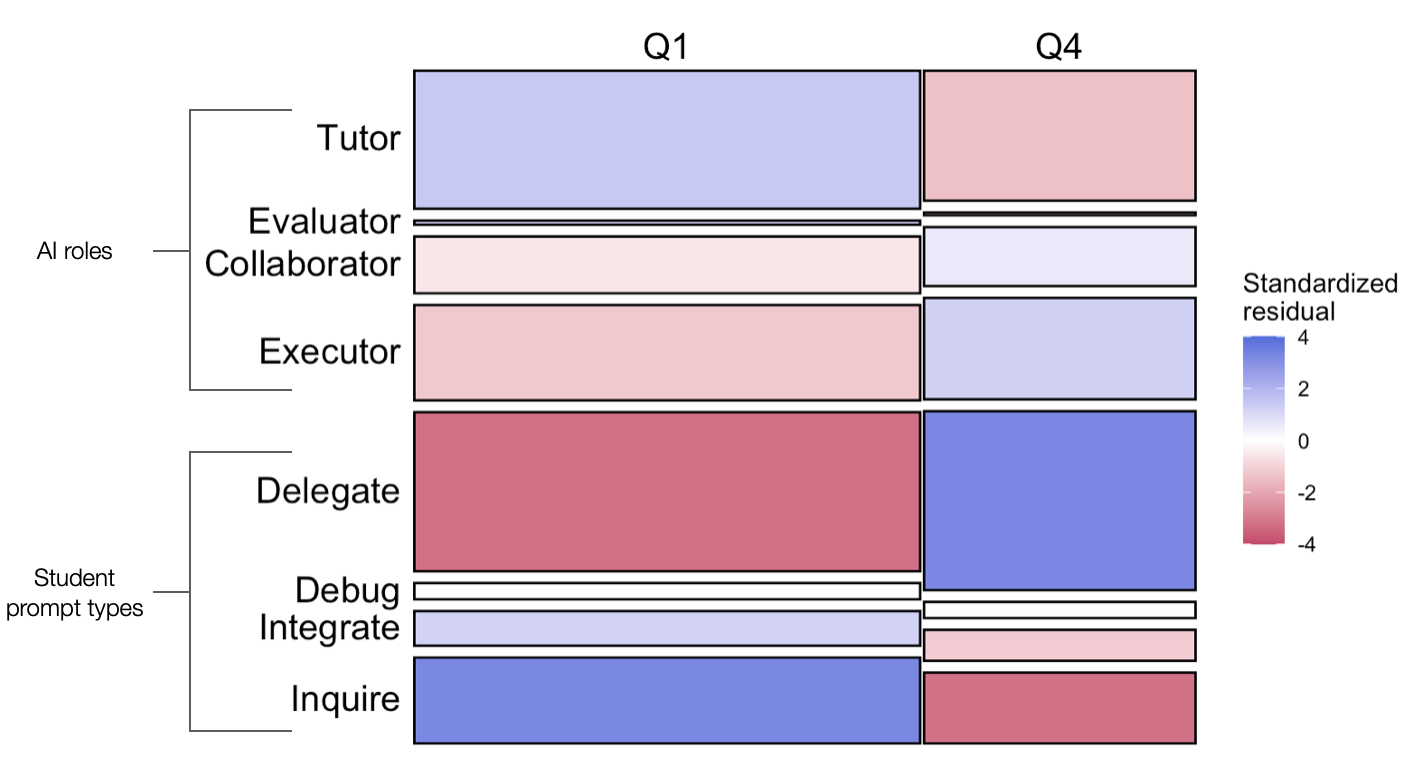}
    \caption{Comparing frequency of individual interaction types between Q1 and Q4 students. Blue shading indicates overrepresentation, while red shading indicates underrepresentation of each element within a group.}
    \label{fig:mosaic_interaction_type}
\end{figure}

To address RQ2, examining how student help-seeking strategies shape AI-enacted roles during programming tasks between Q1 and Q4 students, we first confirmed that the distribution of interaction types differed significantly between groups ($x^2(7) = 22.3, p = 0.002$) using Pearson residual analysis with permutation testing (see Fig. \ref{fig:mosaic_interaction_type}). The most significant difference was observed in the frequency of the \textit{Delegate} code, which was overrepresented among Q4 students and underrepresented among the Q1 students, and the \textit{Inquire} code, which was overrepresented among Q1 students and underrepresented among the Q4 students. Overall, Q1 students received more tutoring from the AI (\textit{Tutor} role), while Q4 students received more ready solutions to their prompts from AI (\textit{Executor} role).

\begin{figure}[!htbp]
    \centering
    \includegraphics[width=0.8\textwidth]{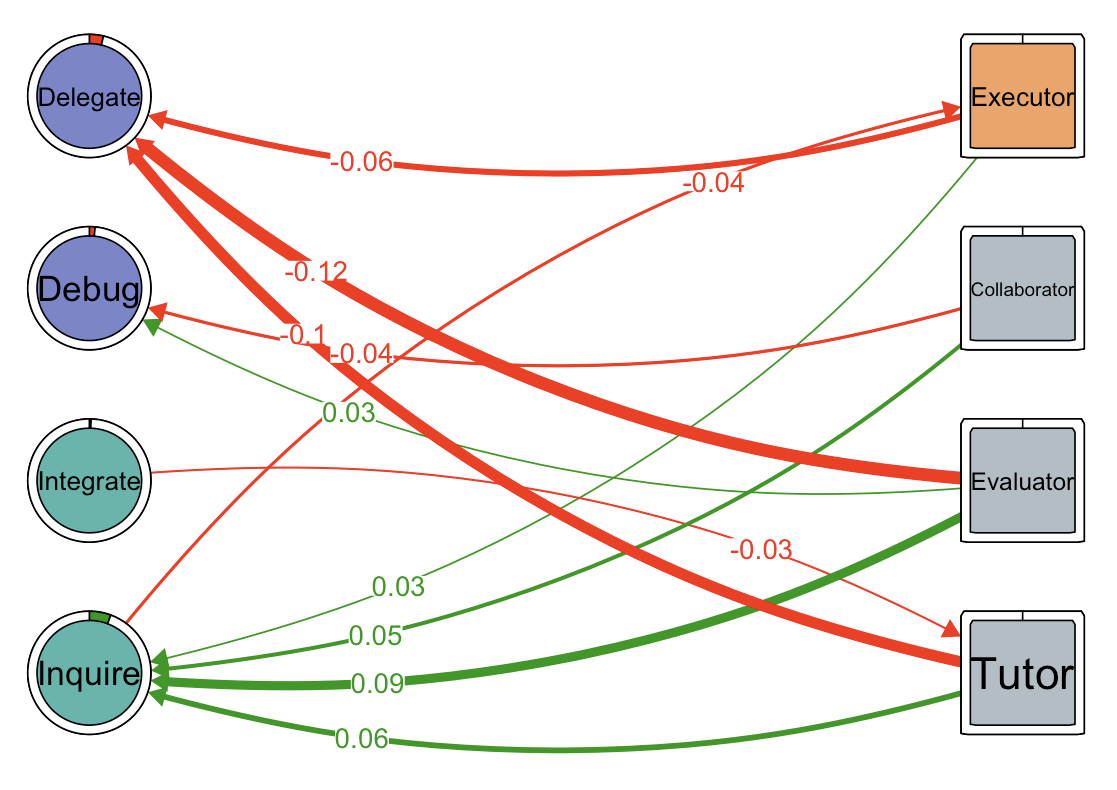}
    \caption{HTNA comparison of interaction types between Q1 and Q4 students. Red edges: stronger for Q1; green edges: stronger for Q4. Blue nodes: instrumental help-seeking; teal nodes: executive help-seeking; gray nodes: AI pedagogical roles; orange: AI non-pedagogical roles. Only transitions over 0.3 are shown.}
    \label{fig:htna_interaction_type}
\end{figure}

The HTNA model indicated that Q4 students tended to \textit{Delegate} tasks (diff: -0.12) to the AI after receiving feedback (\textit{Evaluator}), whereas Q1 students were more likely to \textit{Inquire} further in order to better understand the problem (diff: 0.09) (see Fig. \ref{fig:htna_interaction_type}). A similar contrast was observed following \textit{Tutor} interactions, in which the AI provided conceptual guidance: Q4 students more often transitioned from \textit{Tutor} to \textit{Delegate} (diff: -0.1), while Q1 students typically shifted toward Inquire (diff: 0.06). In addition, the AI was more likely to provide ready-made solutions to Q4 students even when these students demonstrated interest in the subject matter, as reflected by the stronger \textit{Inquire}--\textit{Executor} transition (diff: -0.04). The permutation analysis identified only the \textit{Tutor}--\textit{Delegate} transition as statistically significant (diff: -0.10, $p = 0.007$). Finally, differences were also evident in the initial interaction probabilities: Q1 students more frequently initiated conversations with the AI through instrumental help-seeking, whereas Q4 students were more likely to begin with executive help-seeking.

The most distinguishing interaction type patterns reveal different help-seeking approaches between Q1 and Q4 students (see Table \ref{tab:pattern_interaction_type}). Q1 students demonstrate strong Inquire$\rightarrow$\textit{Tutor} cycles, where inquiry-based help-seeking prompts AI tutoring responses, with this pattern appearing prominently in their top sequences (e.g., Inquire$\rightarrow$\textit{Tutor}$\rightarrow$Inquire: 0.056 vs. 0.036). In contrast, Q4 students show higher proportions of Delegate–\textit{Executor} loops (0.107 vs. 0.088), where task delegation elicits AI execution. They also exhibit more frequent Delegate–\textit{Tutor} patterns (0.071 vs. 0.052), which may indicate solution-seeking behavior rather than inquiry-driven learning. 

\begin{table}[H]
\centering
\scriptsize
\caption{Top 10 pattern differences of interaction types between Q1 and Q4 students, sorted by chi-square statistic. All patterns shown are statistically significant at p < 0.05.}
\label{tab:pattern_interaction_type}
\begin{tabular}{p{7.1cm} p{0.5cm} p{1.3cm} p{0.5cm} p{1.3cm} p{0.7cm}}
\toprule
Pattern & n & Q1 prop. & n & Q4 prop. & stat. \\
\midrule
Delegate$\rightarrow$\textit{Executor}$\rightarrow$Delegate & 413 & 0.088 & 259 & 0.107 & 5.23 \\
\textit{Tutor}$\rightarrow$Inquire$\rightarrow$\textit{Tutor} & 352 & 0.075 & 143 & 0.059 & 5.34 \\
Inquire$\rightarrow$\textit{Tutor}$\rightarrow$Inquire & 261 & 0.056 & 86 & 0.036 & 12.38 \\
Delegate$\rightarrow$\textit{Tutor}$\rightarrow$Delegate & 242 & 0.052 & 171 & 0.071 & 8.96 \\
Inquire$\rightarrow$\textit{Tutor}$\rightarrow$Inquire$\rightarrow$\textit{Tutor} & 224 & 0.051 & 76 & 0.034 & 8.45 \\
\textit{Tutor}$\rightarrow$Inquire$\rightarrow$\textit{Tutor}$\rightarrow$Inquire & 169 & 0.038 & 57 & 0.026 & 6.52 \\
Delegate$\rightarrow$\textit{Executor}$\rightarrow$Delegate$\rightarrow$\textit{Executor}$\rightarrow$Delegate & 168 & 0.041 & 111 & 0.054 & 5.16 \\
Inquire$\rightarrow$\textit{Tutor}$\rightarrow$Inquire$\rightarrow$\textit{Tutor}$\rightarrow$Inquire & 117 & 0.028 & 34 & 0.017 & 6.94 \\
Delegate$\rightarrow$\textit{Collaborator}$\rightarrow$Delegate & 85 & 0.018 & 65 & 0.027 & 5.19 \\
Delegate$\rightarrow$\textit{Tutor}$\rightarrow$Delegate$\rightarrow$\textit{Tutor}$\rightarrow$Delegate & 66 & 0.016 & 53 & 0.026 & 6.44 \\
\bottomrule
\end{tabular}
\end{table}

\subsection{Student-AI Interaction Patterns as Performance Predictors}
Finally, we examined the extent to which student–AI interaction patterns predict student performance (RQ3) by developing two linear regression models using average student grade as the response variable. The interaction element model revealed significant predictive value ($F(10, 99) = 3.85$, $p < 0.001$, adjusted $R^2 = 0.21$). Two interaction elements emerged as significant predictors of student performance. The proportion of \textit{explanation} responses from AI positively predicted grades ($\beta = 0.46$, $SE = 0.15$, $p = 0.003$), suggesting that students who received more conceptual explanations performed better. Conversely, the proportion of assignment elements negatively predicted performance ($\beta = -0.56$, $SE = 0.21$, $p = 0.009$), indicating that students who more frequently pasted assignment text into prompts achieved lower grades. In contrast, the interaction type model failed to yield statistically significant predictors ($F(5, 104) = 0.97$, $p = 0.44$, adjusted $R^2 = -0.001$). None of the help-seeking strategies or AI roles significantly predicted student performance.

\section{Discussion and Conclusion}
Our findings suggest that vibe coding operates as an exercise in intent communication and interpretation, in which the role enacted by AI is shaped by how students frame their help-seeking. Students who engage the AI through inquiry-oriented help-seeking more frequently elicit \textit{Tutor/Collaborator} responses that support exploration and understanding, whereas students who adopt a delegation-oriented stance more often prompt \textit{Executor} responses focused on task completion. This is evident in the recurring interaction patterns that differentiate top-performing (Q1) students from their low-performing (Q4) peers, indicating that differences in interaction style may reinforce different learning trajectories. 
Q4 students’ frequent assignment$\rightarrow$\textit{solution} loops suggest a workflow of pasting tasks and receiving ready-made solutions. Q1 students show exploration-centered, iterative sense-making patterns, often followed by AI feedback or examples. Additionally, AI \textit{explanation} responses emerged as a significant positive predictor of learning outcomes, whereas student prompt \textit{assignment} elements emerged as a significant negative predictor. These results highlight a pedagogical concern regarding vibe coding: student-AI interaction patterns can reinforce unproductive learning behaviors.

Interpreted through help-seeking theory, the observed frequencies and patterns align with the classic distinction between \textit{instrumental} and \textit{executive} help-seeking \cite{karabenick2013help}. Exploration-centered patterns are more consistent with instrumental strategies, where help is used to build understanding. In contrast, repeated assignment$\rightarrow$\textit{solution} loops align with executive help-seeking, where the primary goal is task completion, with limited engagement in sense-making \cite{aleven2003help,karabenick2013help}. These help-seeking moves also function as role invitations: students’ prompts cue the AI to behave as a \textit{Tutor/Collaborator} versus an \textit{Executor}, shaping the resulting collaboration.
These findings have implications for the design of AI systems for programming support in higher education. First, systems should aim to detect and reduce assignment$\rightarrow$\textit{solution} loops. However, the goal should not be to prohibit \textit{Executor} behavior altogether, since direct solutions can be appropriate in some contexts (e.g., to unblock progress, to support time-constrained work). Rather, the goal should be to promote transitions towards a learning-oriented student-AI partnership. For example, after detecting consecutive executive exchanges, a pedagogical assistant could adapt their role from \textit{Executor} into \textit{guiding} behaviors: asking students to share their current attempt, requesting an explanation of their intended approach, offering hints, rather than ready-made solutions. Such adaptations could retain the efficiency benefits of AI while nudging interaction toward instrumental help-seeking, while preserving student agency.

A second design implication concerns the structure of AI responses. Our interaction element analysis, shows that AI responses frequently bundle multiple elements (e.g., \textit{explanation}, \textit{feedback}, \textit{solution}, \textit{instruction}) within a single message, which may reduce opportunities for self-explanation, debugging practice, and critical evaluation \cite{penney2025outcomes,denny2024computing}. The aim should be for the AI responses to invite collaboration. Specifically, assistants could provide layered outputs (\textit{brief diagnosis}$\rightarrow$\textit{hint}$\rightarrow$ (optional) \textit{solution}), require a degree of student self-explanation, or present alternative approaches. Such response structures could increase the likelihood that solutions are integrated through deliberate processing rather than copied with minimal understanding.  

As AI becomes increasingly adept at inferring student intent and generating high-quality solutions with minimal input, the cognitive burden of problem formulation and execution shifts away from the student \cite{fan2025beware}. This redistribution of effort can attenuate learners’ engagement in the problem-solving processes that foster deep learning, effectively relocating both task execution and aspects of learning regulation to the AI system. It is clear that LLMs help level the playing field in many tasks by providing immediate access to effective solutions, even without well-crafted prompts \cite{Acar2023-co}. However, AI assistance may come at the cost of reducing opportunities for students to develop independent problem-solving skills, creativity, and—most importantly—the ability to understand and debug code, which remains central to their training. In other words, the future of software engineering may increasingly demand professionals who can audit and debug AI-generated code rather than those who primarily craft prompts. These shifts call for AI systems that are pedagogically intentional by not just delivering technically sound solutions, but actively supporting learners' capacity to collaborate, critique and problem-solve alongside AI.

%
%
\section*{Acknowledgements}
Kamila Misiejuk conducted this research in cooperation with the LEAD:FUH project funded by the Stiftung Innovation in der Hochschullehre (1001-3223).

\newpage
\bibliographystyle{splncs04}
\bibliography{mybibliography}

@article{tikka2025tna,
  title={tna: An R Package for Transition Network Analysis},
  author={Tikka, Santtu and L{\'o}pez-Pernas, Sonsoles and Saqr, Mohammed},
  journal={Applied Psychological Measurement},
  pages={01466216251348840},
  year={2025},
  publisher={SAGE Publications Sage CA: Los Angeles, CA}
}

@article{mcnichols2025studychat,
  title={The studychat dataset: Student dialogues with chatgpt in an artificial intelligence course},
  author={McNichols, Hunter and Ikram, Fareya and Lan, Andrew},
  journal={arXiv preprint arXiv:2503.07928},
  year={2025}
}

@article{gama2025can,
  title={" Can you feel the vibes?": An exploration of novice programmer engagement with vibe coding},
  author={Gama, Kiev and Calegario, Filipe and Jackson, Victoria and Nolte, Alexander and Morais, Luiz Augusto and Garcia, Vinicius},
  journal={arXiv preprint arXiv:2512.02750},
  year={2025}
}

@inproceedings{staegemann2025did,
  title={How did the Emergence of ChatGPT Impact Stack Overflow? -- A Literature Review},
  author={Staegemann, Daniel and Rizun, Mariia and Haertel, Christian and Pohl, Matthias and Daase, Christian and Turowski, Klaus},
  booktitle={Proceedings of the 33rd International Conference on Information Systems Development (ISD2025)},
  year={2025},
}

@article{shan2025examining,
title={Examining the impact of generative ai on users’ voluntary knowledge contribution: Evidence from a natural experiment on stack overflow},
author={Shan, Guohou and Qiu, Liangfei},
journal={Information Systems Research},
year={2025},
publisher={INFORMS}
}

@inproceedings{saqr2025transition,
  title={Transition network analysis: A novel framework for modeling, visualizing, and identifying the temporal patterns of learners and learning processes},
  author={Saqr, Mohammed and L{\'o}pez-Pernas, Sonsoles and T{\"o}rm{\"a}nen, Tiina and Kaliisa, Rogers and Misiejuk, Kamila and Tikka, Santtu},
  booktitle={Proceedings of the 15th LAK Conference},
  pages={351--361},
  year={2025}
}

@inproceedings{hyrynsalmi2025person,
  title={" Person is a person, a tool is a tool"-ChatGPT’s Role in Student Help-Seeking Behavior and Peer Support},
  author={Hyrynsalmi, Sonja and Tuape, Micheal and Knutas, Antti},
  booktitle={Proceedings of the 33rd ACM FSE Conference},
  pages={783--788},
  year={2025}
}

@article{chen2025unpacking,
  title={Unpacking help-seeking process through multimodal learning analytics: A comparative study of ChatGPT vs Human expert},
  author={Chen, Angxuan and Xiang, Mengtong and Zhou, Junyi and Jia, Jiyou and Shang, Junjie and Li, Xinyu and Ga{\v{s}}evi{\'c}, Dragan and Fan, Yizhou},
  journal={Comput. Educ.},
  volume={226},
  pages={105198},
  year={2025},
  publisher={Elsevier}
}

@ARTICLE{Cubillos2025-tm,
  title     = "Generative artificial intelligence in computer programming: Does
               it enhance learning, motivation, and the learning environment?",
  author    = "Cubillos, Claudio and Mellado, Rafael and Cabrera-Paniagua,
               Daniel and Urra, Enrique",
  journal   = "IEEE Access",
  volume    =  13,
  pages     = "40438--40455",
  year      =  2025
}

@ARTICLE{Garg2025-oy,
  title     = "Enhancing data analysis and programming skills through structured
               prompt training: The impact of generative {AI} in engineering
               education",
  author    = "Garg, Ashish and Nisumba Soodhani, K and Rajendran, Ramkumar",
  journal   = "Comput. Educ.: Artif. Intell.",
  volume    =  8,
  number    =  100380,
  pages     =  100380,
  year      =  2025
}

@INPROCEEDINGS{Lopez-Pernas2025-gt,
  title     = "The dynamics of the self-regulation process in student-{AI}
               interactions: The case of problem-solving in programming
               education",
  author    = "López-Pernas, Sonsoles and Misiejuk, Kamila and Oliveira, Eduardo
               and Saqr, Mohammed",
  booktitle = "Proceedings of the 25th Koli Calling Conference",
  publisher = "ACM",
  pages     = "1--12",
  year      =  2025
}

@INPROCEEDINGS{Amoozadeh2024-od,
  title     = "Student-{AI} Interaction: A Case Study of {CS1} students",
  author    = "Amoozadeh, Matin and Nam, Daye and Prol, Daniel and Alfageeh, Ali
               and Prather, James and Hilton, Michael and Srinivasa Ragavan,
               Sruti and Alipour, Amin",
  booktitle = "Proceedings of the 24th Koli Calling Conference",
  pages     = "1--13",
  year      =  2024
}

@ARTICLE{Nathaniel2025-vl,
  title     = "Literature review on the integration of generative {AI} in
               programming education",
  author    = "Nathaniel, Jemimah and Oyelere, Solomon Sunday and Suhonen,
               Jarkko and Tedre, Matti",
  journal   = "Int. J. Artif. Intell. Educ.",
  volume    =  35,
  number    =  5,
  pages     = "2724--2755",
  year      =  2025
}

@article{karabenick2013help,
  title={Help seeking as a self-regulated learning strategy},
  author={Karabenick, Stuart A and Berger, Jean-Louis},
  journal={Applications of self-regulated learning across diverse disciplines: A tribute to Barry J. Zimmerman},
  pages={237--261},
  year={2013}
}

@article{pintrich2004conceptual,
  title={A conceptual framework for assessing motivation and self-regulated learning in college students},
  author={Pintrich, Paul R},
  journal={Educ. Psychol. Rev.},
  volume={16},
  number={4},
  pages={385--407},
  year={2004}
}

@article{htna2026,
  title={Role Dynamics in Student-AI Collaboration: A Heterogeneous Transition Network Analysis Approach},
  author={López-Pernas, Sonsoles and Misiejuk, Kamila and Tikka, Santtu and Saqr, Mohammed},
  journal={SSRN},
  year={2026},
  doi={10.2139/ssrn.6082190}
}

@article{sun2013mining,
  title={Mining heterogeneous information networks: A structural analysis approach},
  author={Sun, Yizhou and Han, Jiawei},
  journal={ACM SIGKDD Explorations Newsletter},
  volume={14},
  number={2},
  pages={20--28},
  year={2013}
}

@article{aleven2003help,
  title={Help seeking and help design in interactive learning environments},
  author={Aleven, Vincent and Stahl, Elmar and Schworm, Silke and Fischer, Frank and Wallace, Raven},
  journal={Review of Educational Research},
  volume={73},
  number={3},
  pages={277--320},
  year={2003}
}

@inproceedings{ettles2018common,
  title={Common logic errors made by novice programmers},
  author={Ettles, Andrew and Luxton-Reilly, Andrew and Denny, Paul},
  booktitle={Proceedings of the 20th ACE Conference},
  pages={83--89},
  year={2018}
}

@article{denny2024computing,
  title={Computing education in the era of generative AI},
  author={Denny, Paul and Prather, James and Becker, Brett A and Finnie-Ansley, James and Hellas, Arto and Leinonen, Juho and Luxton-Reilly, Andrew and Reeves, Brent N and Santos, Eddie Antonio and Sarsa, Sami},
  journal={Communications of the ACM},
  volume={67},
  number={2},
  pages={56--67},
  year={2024}
}

@inproceedings{kazemitabaar2023studying,
  title={Studying the effect of AI code generators on supporting novice learners in introductory programming},
  author={Kazemitabaar, Majeed and Chow, Justin and Ma, Carl Ka To and Ericson, Barbara J and Weintrop, David and Grossman, Tovi},
  booktitle={Proceedings of the 2023 CHI Conference},
  pages={1--23},
  year={2023}
}

@inproceedings{penney2025outcomes,
  title={Outcomes, perceptions, and interaction strategies of novice programmers studying with ChatGPT},
  author={Penney, Jacob and Acharya, Pawan and Hilbert, Peter and Parekh, Priyanka and Sarma, Anita and Steinmacher, Igor and Gerosa, Marco Aurelio},
  booktitle={Proceedings of the 7th ACM CUI Conference},
  pages={1--15},
  year={2025}
}

@inproceedings{feng2024coprompt,
  title={Coprompt: Supporting prompt sharing and referring in collaborative natural language programming},
  author={Feng, Li and Yen, Ryan and You, Yuzhe and Fan, Mingming and Zhao, Jian and Lu, Zhicong},
  booktitle={Proceedings of the 2024 CHI Conference},
  pages={1--21},
  year={2024}
}

@article{fan2025beware,
  title={Beware of metacognitive laziness: Effects of generative artificial intelligence on learning motivation, processes, and performance},
  author={Fan, Yizhou and Tang, Luzhen and Le, Huixiao and Shen, Kejie and Tan, Shufang and Zhao, Yueying and Shen, Yuan and Li, Xinyu and Ga{\v{s}}evi{\'c}, Dragan},
  journal={BJET},
  volume={56},
  number={2},
  pages={489--530},
  year={2025},
  publisher={Wiley Online Library}
}

@inproceedings{prather2024widening,
  title={The widening gap: The benefits and harms of generative ai for novice programmers},
  author={Prather, James and Reeves, Brent N and Leinonen, Juho and MacNeil, Stephen and Randrianasolo, Arisoa S and Becker, Brett A and Kimmel, Bailey and Wright, Jared and Briggs, Ben},
  booktitle={Proceedings of the 2024 ACM Conference on International Computing Education Research-Volume 1},
  pages={469--486},
  year={2024}
}

@article{fan2025impact,
  title={The impact of AI-assisted pair programming on student motivation, programming anxiety, collaborative learning, and programming performance: a comparative study with traditional pair programming and individual approaches},
  author={Fan, Guangrui and Liu, Dandan and Zhang, Rui and Pan, Lihu},
  journal={International Journal of STEM Education},
  volume={12},
  number={1},
  pages={16},
  year={2025},
  publisher={Springer}
}

@article{wang2023matters,
  title={What matters in AI-supported learning: A study of human-AI interactions in language learning using cluster analysis and epistemic network analysis},
  author={Wang, Xinghua and Liu, Qian and Pang, Hui and Tan, Seng Chee and Lei, Jun and Wallace, Matthew P and Li, Linlin},
  journal={Computers \& Education},
  volume={194},
  pages={104703},
  year={2023},
  publisher={Elsevier}
}

@article{mailach2025ok,
  title={“Ok Pal, we have to code that now”: interaction patterns of programming beginners with a conversational chatbot},
  author={Mailach, Alina and Gorgosch, Dominik and Siegmund, Norbert and Siegmund, Janet},
  journal={Empirical Software Engineering},
  volume={30},
  number={1},
  pages={34},
  year={2025},
  publisher={Springer}
}

@ARTICLE{Acar2023-co,
  title    = "{AI} Prompt Engineering Isn’t the Future",
  author   = "Acar, Oguz A",
  journal  = "Harvard Business Review",
  month    =  jun,
  year     =  2023,
  url      = "https://hbr.org/2023/06/ai-prompt-engineering-isnt-the-future",
  issn     = "0017-8012",
  language = "en"
}

\newpage
\appendix
\section*{Appendix}
\addcontentsline{toc}{section}{Appendix}
Supplementary figures, tables, and analyses
\vspace{1em}
 
\section{Interaction elements analysis across all performance level groups}\label{app:elements}
\begin{figure}[!htbp]
\centering
    \includegraphics[width=0.8\textwidth]{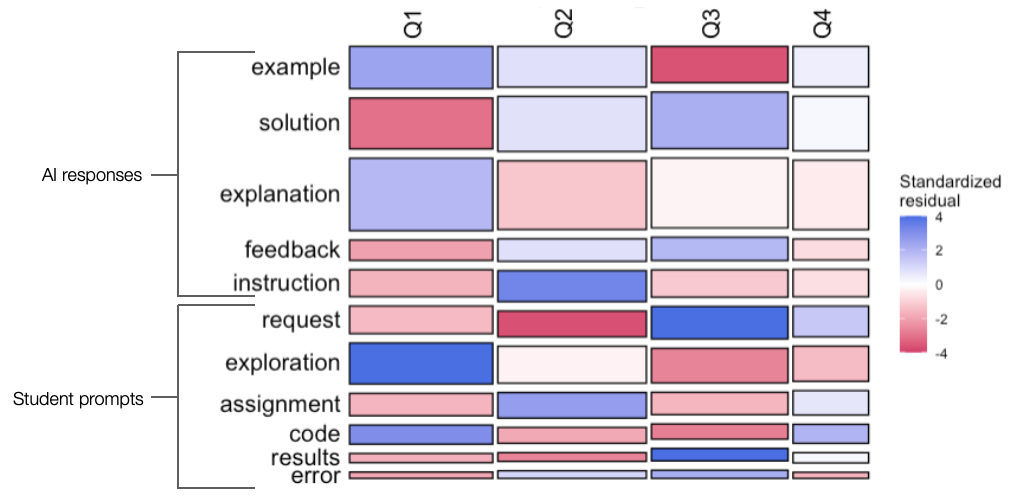}
    \caption{Pearson residual analysis comparing the frequency of individual interaction elements across all performance level groups ($x^2(30) = 137.5, p < 0.001$). Blue shading indicates overrepresentation, while red shading indicates underrepresentation of each element within a group.}
\end{figure}
 
\begin{figure}[!htbp]
    \includegraphics[width=0.9\linewidth]
    {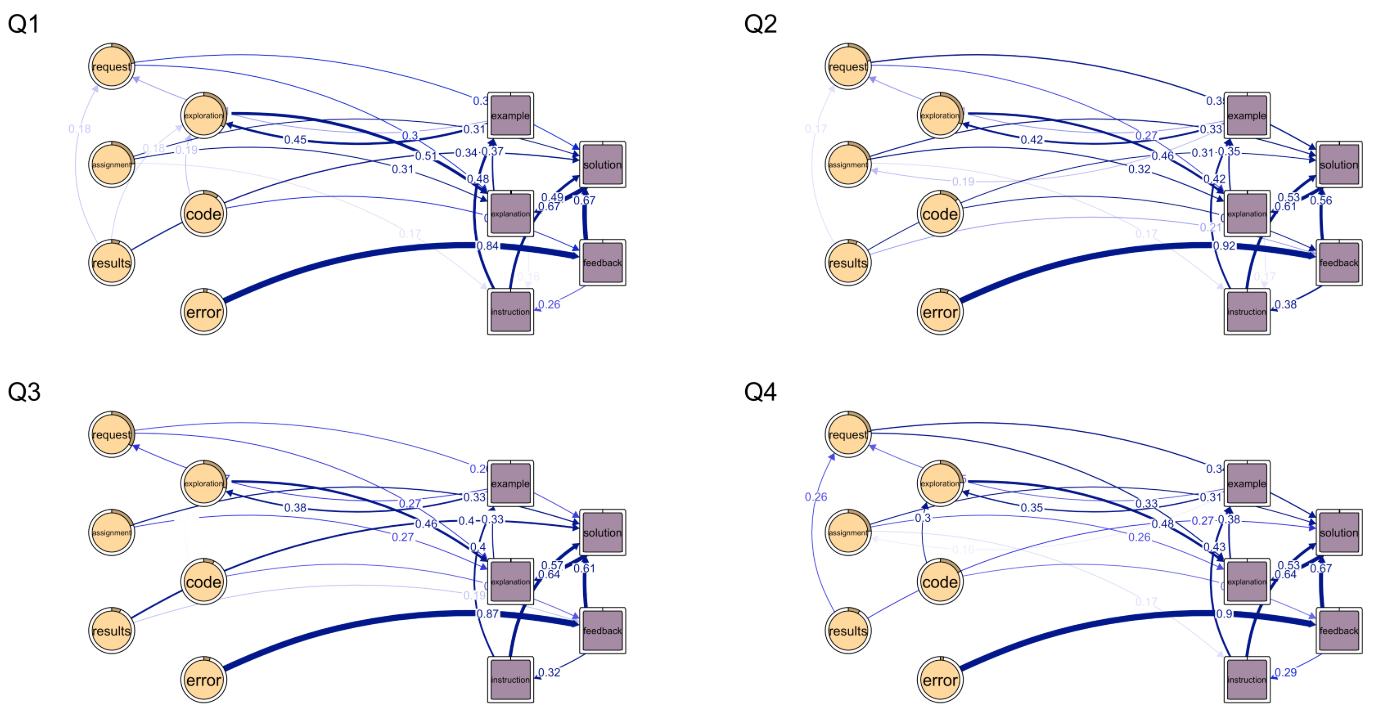}
    \caption{Heterogeneous Temporal Network Analysis (HTNA) models of interaction patterns for each performance groups. Directed edges show transition probabilities between patterns, with edge width proportional to transition strength. Pie rings indicate the starting probability for each interaction pattern.}
\end{figure}
 
\clearpage
\begin{table}[!htbp]
\centering
\caption{Top 10 interaction element patterns showing significant frequency differences across all performance levels (all $p<0.001$). Student prompt elements appear in normal font; AI response elements appear in italics.}
\label{tab:top10-elements}
\resizebox{\textwidth}{!}{%
\begin{tabular}{p{8cm} r r r r r r r r r}
\hline
 & Q1 & Q1 & Q2 & Q2 & Q3 & Q3 & Q4 & Q4 & \\
Pattern & n & Prop. & n & Prop. & n & Prop. & n & Prop. & Stat.\\
\hline
exploration$\rightarrow$\textit{explanation}$\rightarrow$\textit{example}$\rightarrow$exploration & 186 & 0.023 & 154 & 0.018 & 115 & 0.015 & 55 & 0.013 & 20.41 \\
code$\rightarrow$exploration$\rightarrow$\textit{explanation} & 39 & 0.005 & 17 & 0.002 & 13 & 0.002 & 25 & 0.006 & 24.85 \\
\textit{feedback}$\rightarrow$instruction$\rightarrow$\textit{solution} & 143 & 0.017 & 238 & 0.027 & 195 & 0.024 & 87 & 0.020 & 21.78 \\
assignment$\rightarrow$\textit{solution}$\rightarrow$assignment & 17 & 0.002 & 32 & 0.004 & 21 & 0.003 & 33 & 0.007 & 27.68 \\
assignment$\rightarrow$\textit{solution}$\rightarrow$assignment$\rightarrow$\textit{solution} & 10 & 0.001 & 14 & 0.002 & 14 & 0.002 & 24 & 0.006 & 29.37 \\
\textit{solution}$\rightarrow$\textit{explanation}$\rightarrow$code$\rightarrow$exploration & 22 & 0.003 & 13 & 0.002 & 21 & 0.003 & 25 & 0.006 & 19.74 \\
code$\rightarrow$exploration$\rightarrow$\textit{explanation}$\rightarrow$\textit{example} & 19 & 0.002 & 9 & 0.001 & 6 & 0.001 & 17 & 0.004 & 20.55 \\
code$\rightarrow$exploration$\rightarrow$\textit{feedback}$\rightarrow$\textit{solution} & 18 & 0.002 & 17 & 0.002 & 21 & 0.003 & 26 & 0.006 & 19.83 \\
request$\rightarrow$\textit{explanation}$\rightarrow$\textit{example}$\rightarrow$code & 12 & 0.001 & 9 & 0.001 & 14 & 0.002 & 19 & 0.004 & 19.44 \\
\textit{explanation}$\rightarrow$request$\rightarrow$instruction & 12 & 0.001 & 8 & 0.001 & 31 & 0.004 & 14 & 0.003 & 20.89 \\
\hline
\end{tabular}%
}
\end{table}
 
\begin{table}[!htbp]
\centering
\caption{Linear regression model of interaction elements with grade as dependent variable}
\begin{threeparttable}
\begin{tabular}{lcccc}
\toprule
Variable & Estimate & Std. Error & $t$ value & $p$ value \\
\midrule
(Intercept) & 0.8767 & 0.0376 & 23.33 & $<0.001^{***}$ \\
Error & 0.4889 & 0.6268 & 0.78 & 0.437 \\
Code & $-0.0131$ & 0.2576 & $-0.05$ & 0.960 \\
Exploration & $-0.1634$ & 0.1850 & $-0.88$ & 0.379 \\
Results & 0.4726 & 0.2391 & 1.98 & $0.051^{.}$ \\
Assignment & $-0.5575$ & 0.2085 & $-2.67$ & $0.009^{**}$ \\
Request & 0.0129 & 0.1838 & 0.07 & 0.944 \\
Instruction & 0.0561 & 0.2447 & 0.23 & 0.819 \\
Feedback & $-0.3076$ & 0.3013 & $-1.02$ & 0.310 \\
Explanation & 0.4615 & 0.1505 & 3.07 & $0.003^{**}$ \\
Solution & 0.0835 & 0.1608 & 0.52 & 0.605 \\
\bottomrule
\end{tabular}
\begin{tablenotes}
\small
\item Residual standard error: 0.0579 on 99 degrees of freedom
\item Multiple $R^2$: 0.28, Adjusted $R^2$: 0.207
\item $F$-statistic: 3.85 on 10 and 99 DF, $p$-value: 0.0002
\item Significance codes: $^{***}p<0.001$, $^{**}p<0.01$, $^{*}p<0.05$, $^{.}p<0.1$
\item VIF values: Error = 2.26, Code = 1.66, Exploration = 3.11, Results = 1.86, Assignment = 3.13, Request = 2.79, Instruction = 1.84, Feedback = 2.75, Explanation = 2.24, Solution = 2.56
\end{tablenotes}
\end{threeparttable}
\end{table}

\clearpage
\section{Interaction elements types across all performance level groups}\label{app:types}
 
\begin{figure}[!htbp]
\centering
    \includegraphics[width=0.6\linewidth]{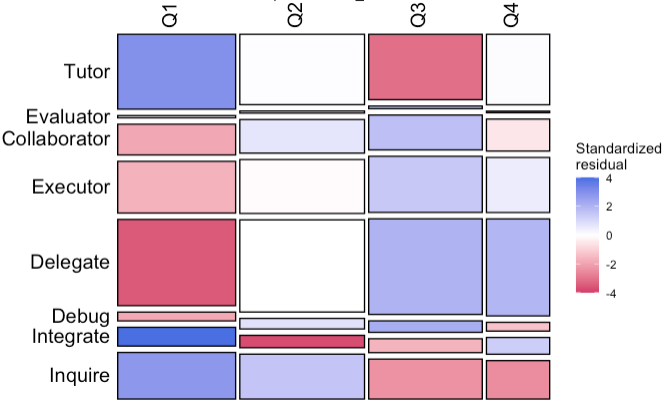}
    \caption{Pearson residual analysis comparing the frequency of individual interaction types across all performance level groups ($x^2(21) = 80.8, p < 0.001$). Blue shading indicates overrepresentation, while red shading indicates underrepresentation of each element within a group.}
\end{figure}
 
\begin{figure}[!htbp]
\includegraphics[width=1.0\linewidth]{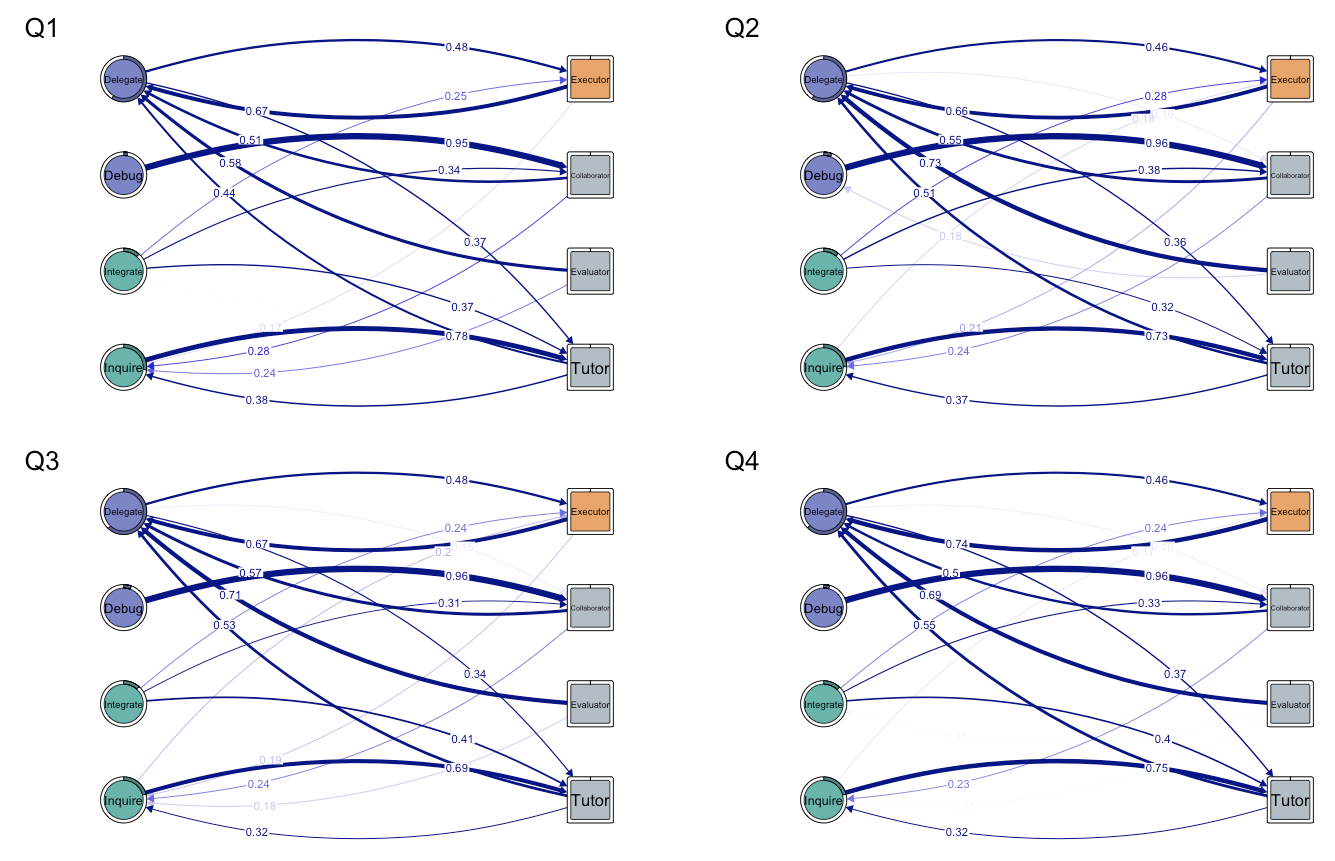}
    \caption{Heterogeneous Temporal Network Analysis (HTNA) models of interaction types for each performance groups. Directed edges show transition probabilities between patterns, with edge width proportional to transition strength. Pie rings indicate the starting probability for each interaction pattern.}
\end{figure}
\newpage
\begin{table}[h]
\centering
\caption{Top 10 interaction type patterns showing significant frequency differences across all performance levels (all $p<0.001$). Student prompt elements appear in normal font; AI response elements appear in italics.}
\label{tab:top10-types}
\resizebox{\textwidth}{!}{%
\begin{tabular}{p{7cm} r r r r r r r r r}
\hline
 & Q1 & Q1 & Q2 & Q2 & Q3 & Q3 & Q4 & Q4 & \\
Pattern & n & Prop. & n & Prop. & n & Prop. & n & Prop. & Stat.\\
\hline
\textit{Tutor}$\rightarrow$Inquire$\rightarrow$\textit{Tutor} & 352 & 0.075 & 325 & 0.066 & 237 & 0.053 & 143 & 0.059 & 17.63 \\
Inquire$\rightarrow$\textit{Tutor}$\rightarrow$Inquire$\rightarrow$\textit{Tutor} & 224 & 0.051 & 208 & 0.045 & 134 & 0.032 & 76 & 0.034 & 21.86 \\
Inquire$\rightarrow$\textit{Tutor}$\rightarrow$Inquire & 261 & 0.056 & 261 & 0.053 & 152 & 0.034 & 86 & 0.036 & 33.04 \\
\textit{Tutor}$\rightarrow$Inquire$\rightarrow$\textit{Tutor}$\rightarrow$Inquire & 169 & 0.038 & 158 & 0.034 & 93 & 0.022 & 57 & 0.026 & 21.31 \\
\textit{Tutor}$\rightarrow$Inquire$\rightarrow$\textit{Tutor}$\rightarrow$Inquire$\rightarrow$\textit{Tutor} & 149 & 0.036 & 131 & 0.030 & 83 & 0.021 & 50 & 0.025 & 16.87 \\
Inquire$\rightarrow$\textit{Tutor}$\rightarrow$Inquire$\rightarrow$\textit{Tutor}$\rightarrow$Inquire & 117 & 0.028 & 115 & 0.027 & 55 & 0.014 & 34 & 0.017 & 24.88 \\
Inquire$\rightarrow$\textit{Tutor}$\rightarrow$Integrate & 65 & 0.014 & 29 & 0.006 & 39 & 0.009 & 27 & 0.011 & 16.73 \\
\textit{Tutor}$\rightarrow$Integrate$\rightarrow$Collaborator & 49 & 0.010 & 18 & 0.004 & 27 & 0.006 & 15 & 0.006 & 17.27 \\
Collaborator$\rightarrow$\textit{Delegate}$\rightarrow$Collaborator$\rightarrow$\textit{Delegate} & 20 & 0.005 & 35 & 0.008 & 54 & 0.013 & 11 & 0.005 & 22.10 \\
Inquire$\rightarrow$\textit{Tutor}$\rightarrow$Inquire$\rightarrow$\textit{Executor} & 19 & 0.004 & 37 & 0.008 & 12 & 0.003 & 6 & 0.003 & 15.35 \\
\hline
\end{tabular}%
}
\end{table}

\begin{table}[!htbp]
\centering
\caption{Linear regression model of interaction types with grade as dependent variable}
\begin{threeparttable}
\begin{tabular}{lcccc}
\toprule
Variable & Estimate & Std. Error & $t$ value & $p$ value \\
\midrule
(Intercept) & 0.9038 & 0.0092 & 97.76 & $<0.001^{***}$ \\
Debug & $-0.0004$ & 0.0012 & $-0.38$ & 0.710 \\
Inquire & 0.0001 & 0.0002 & 0.24 & 0.810 \\
Integrate & 0.0007 & 0.0008 & 0.79 & 0.430 \\
Evaluator & $-0.0005$ & 0.0034 & $-0.15$ & 0.880 \\
Executor & 0.0004 & 0.0004 & 1.04 & 0.300 \\
\bottomrule
\end{tabular}
\begin{tablenotes}
\small
\item Residual standard error: 0.065 on 104 degrees of freedom
\item Multiple $R^2$: 0.0445, Adjusted $R^2$: $-0.00139$
\item $F$-statistic: 0.97 on 5 and 104 DF, $p$-value: 0.44
\item Significance codes: $^{***}p<0.001$, $^{**}p<0.01$, $^{*}p<0.05$, $^{.}p<0.1$
\item VIF values: Debug = 2.29, Inquire = 1.77, Integrate = 1.92, Evaluator = 1.52, Executor = 2.68
\end{tablenotes}
\end{threeparttable}
\end{table}

\end{document}